\documentclass[10pt,twocolumn,letterpaper]{article}

\usepackage[pagenumbers]{cvpr} 

\definecolor{cvprblue}{rgb}{0.21,0.49,0.74}
\usepackage[pagebackref,breaklinks,colorlinks,allcolors=cvprblue]{hyperref}
\usepackage{xcolor} 
\usepackage{algorithm}
\usepackage[noend]{algpseudocode}
\usepackage{multirow}
\usepackage{subfiles}

\usepackage{color}
\usepackage{pifont}
\usepackage{xcolor}
\usepackage{colortbl}

\usepackage{arydshln}

\usepackage{caption}

\title{MambaScope: Coarse-to-Fine Scoping for Efficient Vision Mamba}

\author{Shanhui Liu\\
The University of Sydney\\
{\tt\small sliu0390@uni.sydney.edu.au}
\and
Rui Xu\\
Wuhan University\\
{\tt\small xurui7943@gmail.com}
\and
Yunke Wang\\
The University of Sydney\\
{\tt\small yunke.wang@sydney.edu.au}
}

\begin{document}
\maketitle
\begin{abstract}
Vision Mamba has emerged as a promising and efficient alternative to Vision Transformers, yet its efficiency remains fundamentally constrained by the number of input tokens. Existing token reduction approaches typically adopt token pruning or merging to reduce computation. However, they inherently lead to information loss, as they discard or compress
token representations. This problem is exacerbated when applied uniformly to fine-grained token representations across all images, regardless of visual complexity. We observe that not all inputs require fine-grained processing. Simple images can be effectively handled at coarse resolution, while only complex ones may warrant refinement. Based on this insight, we propose MambaScope, an adaptive framework for efficient inference for vision Mamba. MambaScope first performs coarse-grained inference by dividing the input image into large patches, significantly reducing the token length and computation. When the model’s prediction confidence is low, selected regions are re-processed at a finer resolution to recover critical visual details with minimal additional cost. This dynamic resolution assignment strategy allows MambaScope to allocate computation adaptively according to image complexity, ensuring efficient processing without compromising essential visual information. Experiments on various vision tasks demonstrate that MambaScope outperforms both the baseline Vision Mamba and state-of-the-art token reduction techniques in terms of accuracy and efficiency.
\end{abstract}    
\section{Introduction}
Vision Mamba~\cite{vim,vim4,mambasurvy} has recently emerged as a promising alternative to Vision Transformers~\cite{attn,dosovitskiy2020vit} for vision tasks. By leveraging state-space models (SSMs)~\cite{ssm,mamba} to replace self-attention, it achieves linear-time sequence modeling while maintaining strong performance. This architectural advantage makes Vision Mamba particularly attractive for efficiency-critical applications. However, its overall computational cost still scales with the number of input tokens. As a result, Vision Mamba's efficiency remains fundamentally constrained by token length, especially in scenarios requiring real-time inference or deployment on resource-limited devices.

\begin{figure}[!t]
    \centering
    \includegraphics[width=0.99\linewidth]{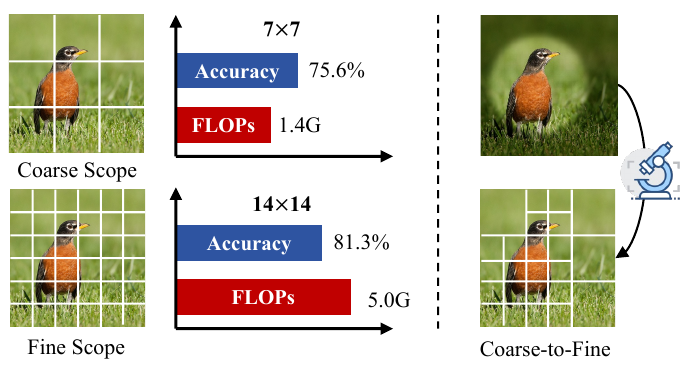}
    \caption{Comparison of coarse scope (7$\times$7), fine scope (14$\times$14) and \textbf{coarse-to-fine scope} in Vision Mamba. For simplicity, we use less patches here for better visualization. While fine-grained processing yields higher accuracy, it incurs significantly higher computational cost. In contrast, coarse-to-fine approach selectively applies fine-grained analysis, achieving a better accuracy-efficiency trade-off by focusing computation on informative regions.}
    \vskip -0.1in
    \label{fig:motivation}
\end{figure}

Spatial redundancy, which has been extensively studied in Vision Transformers (ViTs)~\cite{han2021dynamicnn, liang2022evit, meng2021adavit, song2023dge, wang2021dynamic}, is also likely to exist in Vision Mamba models. This redundancy typically manifests at the token level, where representing an image with an excessive number of visual tokens~\cite{liu2023revisiting} leads to increased computational overhead and degraded inference efficiency. To alleviate the token-level computational burden, several prior works have explored token reduction strategies~\cite{vimpruning1, vimpruning2, vimmerge} such as token pruning and token merging. These techniques aim to reduce the number of tokens processed by the network, thereby lowering computational cost. However, these methods inherently lead to information loss, as they either discard potentially useful tokens or collapse multiple token representations into a single one. This issue becomes more severe when token reduction is performed after converting all images into fine-grained token representations. Such approaches process all inputs with equal granularity, regardless of their visual complexity, which can result in unnecessary information loss. For instance, pruning fine-grained tokens from visually simple regions may discard information that would not have been lost under a coarser representation.

A key observation motivating our work is that not all visual inputs require fine-grained, token-level processing. In fact, coarse-grained patches alone can already deliver strong performance. As shown in Figure~\ref{fig:motivation}, using only \(7 \times 7\) coarse tokens achieves a Top-1 accuracy of 75.6\% with just 1.4 GFLOPs and is substantially more efficient than the full-resolution \(14 \times 14\) configuration, which yields 81.3\% accuracy at a cost of 5.0 GFLOPs. This result underscores the potential of adaptive token processing. In fact, some images such as those with simple structures or clearly localized content can be effectively handled using coarse-grained representations. In contrast, others with more complex textures or spatial layouts may benefit from higher-resolution analysis. This suggests that applying a uniform level of granularity to all inputs is computationally inefficient. 


Based on this insight, we propose \textit{Coarse-to-Fine Scoping for Efficient Vision Mamba (MambaScope)}, a novel adaptive framework for efficient inference. MambaScope adopts a coarse-to-fine scoping strategy for dynamic resolution assignment. It first performs coarse-grained inference by dividing the input image into large patches, which significantly reduces token length and computation. When the model's initial prediction yields low confidence, MambaScope selectively identifies informative regions that require further refinement and reprocesses them at a finer resolution. This allows the model to recover crucial visual details only when necessary, avoiding the overhead of applying fine-grained processing to all images regardless of their complexity. Importantly, it preserves essential information. Unlike previous approaches that statically reduce fine-grained tokens across all images, MambaScope dynamically allocates computation based on the visual complexity of each image, striking a better balance between efficiency and task performance.


Our contributions can be summarized as follows:
\begin{itemize}
    \item We propose Coarse-to-Fine Scoping for Efficient Vision Mamba (MambaScope), an adaptive framework for efficient inference that dynamically adjusts token resolution according to input complexity. This strategy reduces overall token usage while maintaining essential visual information.

    \item We leverage the Mamba architecture’s rich channel-wise representations to effectively identify informative tokens that merit finer-grained processing, without introducing additional supervision mechanisms.
    
    \item We introduce a token scanning mechanism to integrate coarse and fine-resolution tokens, ensuring coherent cross-scale information flow. We further incorporate coarse-stage features into the second-stage encoding to enhance contextual representation during inference.
\end{itemize}

Extensive experiments on various vision tasks (\textit{e.g.,} classification, segmentation, and detection) validate the effectiveness of our MambaScope. For instance, compared to the original ViM, MambaScope reduces FLOPs by 47\% while preserving comparable accuracy on ImageNet. Moreover, MambaScope outperforms existing state-of-the-art methods under the same computational budget.

\begin{figure*}[!t]
  \centering
    \includegraphics[width=0.98\linewidth]{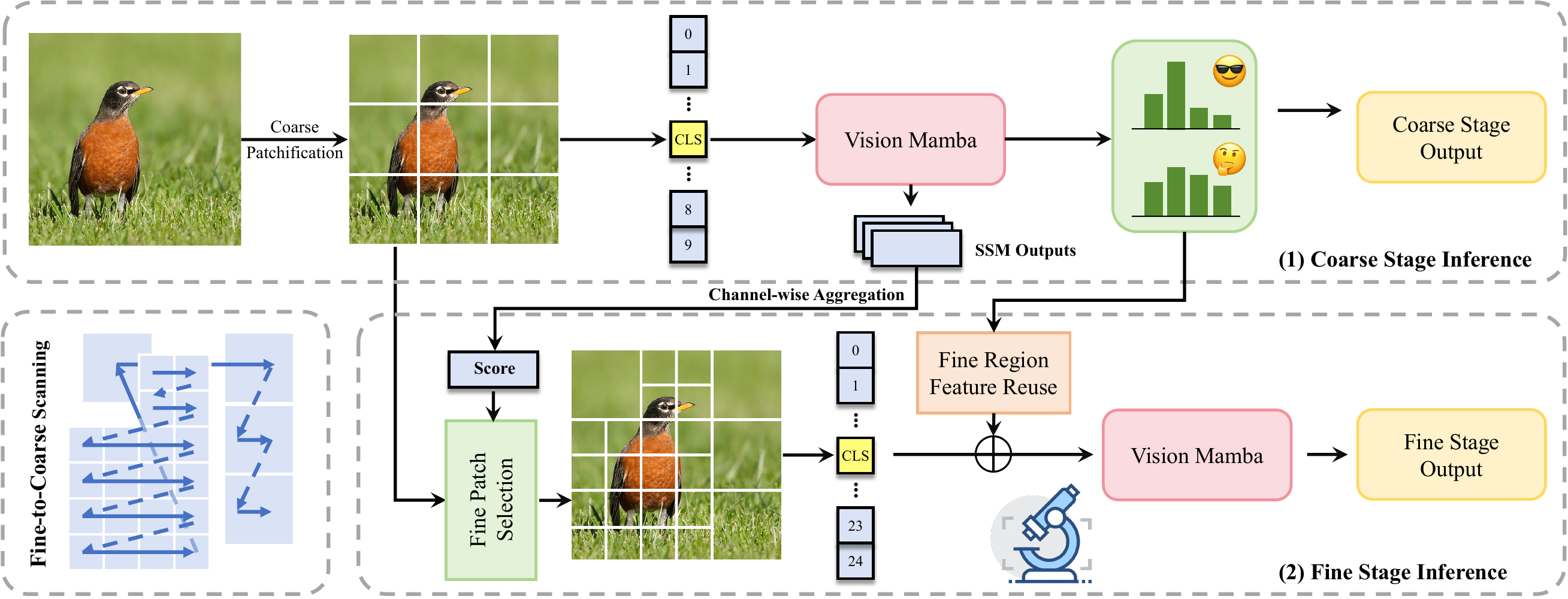}
  \caption{Inference pipeline of MambaScope. The input image is first processed in the coarse stage, using large patches to produce an initial output. If the model's confidence is insufficient, it proceeds to the fine stage, where it identifies informative regions, splits them into smaller patches, and reuses coarse-stage features to refine the output. Both stages share the same network parameters.
  }
  \label{fig:structure}
\end{figure*}
\section{Related Work}

\textbf{Mamba-based Vision Architectures.}  
The success of Mamba in NLP has inspired its adaptation to vision tasks, addressing the challenge of modeling 2D spatial dependencies with 1D sequence models. Building on the SSM~\cite{mehta2022gss, smith2022s4d} foundation of Mamba~\cite{mamba}, a series of follow-up studies~\cite{othermamba1, huang2024localmamba, othermamba3} have achieved state-of-the-art results on various vision benchmarks by employing Mamba-based backbones. VMamba~\cite{vmamba} introduces the SS2D module with multi-directional scanning to capture 2D structures, while PlainMamba~\cite{plainmamba} employs zigzag scanning and direction-aware updates for improved feature fusion. LocalMamba~\cite{huang2024localmamba} enhances local interactions via state space layers, reducing reliance on self-attention. QuadMamba~\cite{xie2024quadmamba} adopts a quadrilateral traversal for directional encoding. ViM~\cite{vim} integrates Mamba~\cite{mamba,ssm2} blocks with positional embeddings, showing strong performance on vision benchmarks. These works demonstrate the growing potential of Mamba-based SSMs in efficient and scalable visual representation learning.

\noindent\textbf{Token Compression in Vision Backbones.}  
Several token compression methods have been proposed for Vision Transformers (ViTs). Some approaches~\cite{rao2021dynamicvit, pan2021iared2, liang2022evit, kong2022spvit} dynamically prune unimportant tokens during inference to improve computational efficiency. In contrast, Evo-ViT~\cite{xu2022evovit} retains all tokens but allocates fewer computational resources to the less informative ones. DVT~\cite{wang2021dynamic} determines the number of tokens per input by cascading three transformers, accelerating inference at the expense of increased storage. PS-ViT (T2T)~\cite{tvt} adopts a top-down paradigm to discard less informative patches in a hierarchical manner. Similarly, CF-ViT~\cite{cfvit} follows a coarse-to-fine strategy that first performs inference on low-resolution patches and progressively refines salient regions, achieving adaptive computational allocation across different image complexities. PatchMerger~\cite{renggli2022mergetokens} utilizes spatial attention to aggregate input tokens into a compact set. ToMe~\cite{bolya2023tokmerging} merges similar tokens. However, due to fundamental architectural differences, these methods are not directly applicable to Vision Mamba~\cite{vim}. Several attempts have been made to enable token compression within Mamba-based models. HiddenAlign (HA)~\cite{zhan2024tokenpruning}, DyVM~\cite{wang2021dynamic}, and RMeeto~\cite{shi2024fastmamba} explore token-level pruning in Vision Mamba; however, these approaches still lead to varying degrees of information loss. In contrast, MambaScope dynamically allocates computational resources based on the visual complexity of each image, achieving a more favorable trade-off between efficiency and task performance.


\section{Preliminary}

State Space Models (SSMs)~\cite{ssm, ssm2, ssm4} are continuous systems that map an input sequence \( x(t) \in \mathbb{R}^L \) to an output sequence \( y(t) \in \mathbb{R}^L \) via a hidden state \( h(t) \in \mathbb{R}^N \):
\begin{equation}
\frac{d}{dt} h(t) = A h(t) + B x(t), \quad y(t) = C h(t),
\end{equation}
where \( A \in \mathbb{R}^{N \times N} \), \( B \in \mathbb{R}^{N \times L} \), and \( C \in \mathbb{R}^{L \times N} \) are learnable system parameters.

Mamba~\cite{mamba} discretizes the continuous system using the zero-order hold (ZOH) with a timescale parameter \( \Delta \in \mathbb{R} \):
\begin{equation}
\bar{A} = \exp(\Delta A), \quad \bar{B} = (\Delta A)^{-1} (\exp(\Delta A) - I) \Delta B,
\end{equation}
and the discrete version of Eq.~(1) becomes:
\begin{equation}
h_t = \bar{A} h_{t-1} + \bar{B} x_t, \quad y_t = C h_t.
\end{equation}
The recurrent formulation in Eq.~(3) can be reformulated and computed as a global convolution:
\begin{equation}
y = x * \bar{K}, \quad \bar{K} = (C \bar{B},\, C \bar{A} \bar{B},\, \dots,\, C \bar{A}^{L-1} \bar{B}),
\end{equation}
where \( \bar{K} \in \mathbb{R}^L \) is the convolution kernel and \( * \) denotes 1D convolution. 


Although Mamba scales linearly with sequence length, adapting it to 2D images typically involves various token traversal strategies, and the large number of visual tokens still leads to considerable computational overhead.

\section{Methodology}

In this section, we propose a new framework called \textbf{MambaScope}, which is designed to reduce the computational cost of Vision Mamba at the token level. As illustrated in Fig.~\ref{fig:structure}, MambaScope introduces a two-stage dynamic inference process.
The coarse inference stage is first performed using a short token sequence for prediction. If the model is uncertain to its prediction, the most informative regions of the image will be subdivided further for fine-grained predication. The details are elaborated in the following subsections.

\subsection{Inference with Coarse Scoping}

As the computational cost of Vision Mamba increases as the length of the input sequence scales, MambaScope encodes the input image using a low-resolution patch representation at the coarse stage. This coarse-level processing is sufficient for most samples with clear visual cues. The coarse-stage input sequence is defined as $\mathcal{S}_0 = [s_0^{(0)}; s_1^{(0)}; \dots; s_{m}^{(0)}] + \mathcal{P}_0$,
where $m$ denotes the number of coarse tokens and $\mathcal{P}_0$ is the corresponding positional embedding. Note that the class token $s_{\text{cls}}$ is inserted in the middle of the sequence. This sequence is processed by a Vision Mamba~\cite{vim} encoder $\Phi$ composed of $d$ layers:
\begin{equation}
\Phi(\mathcal{S}_0) = [s_0^{(d)}; s_1^{(d)}; \dots; s_{m}^{(d)}].
\end{equation}
The class token $s_{\text{cls}}^{(d)}$, acting as the global summary representation, is then forwarded to a classification head $F$ to produce a probability distribution over $C$ categories:
\begin{equation}
\mathbf{q} = F(s_{\text{cls}}^{(d)}) = [q_1, q_2, \dots, q_C].
\end{equation}
The predicted label is obtained by:
\begin{equation}
\hat{c} = \arg\max_{c} q_c,
\end{equation}
with the associated confidence score given by $q_{\hat{c}}$. If confidence exceeds a predefined threshold $\eta$, (\textit{i.e.}, $q_{\hat{c}} \ge \eta)$, the prediction is accepted, and class $\hat{c}$ is returned. Otherwise, the model proceeds to a fine-grained stage for further processing of informative high-resolution patches.

\subsection{Coarse-to-Fine Scoping}
A naive solution to improve model performance is to subdivide all coarse-level patches into smaller ones. However, this substantially increases the number of tokens and the associated computational cost. To mitigate this, we propose a fine patch selection strategy that selectively refines only the most informative regions that are most relevant to the final prediction. Consequently, identifying these informative patches becomes a critical factor for efficient refinement.

\subsubsection{Token Importance Score in Vision Mamba} 
In vision Mamba, each layer $\ell$ receives an input token sequence $T_{\ell-1} \in \mathbb{R}^{B \times N \times D}$, where $B$ is the batch size, $N$ the number of tokens, and $D$ the feature dimension. This sequence is first projected into a lower-dimensional space:
\begin{equation}
X' \in \mathbb{R}^{B \times N \times D'}.
\end{equation}

To model long-range dependencies, $X'$ is processed in both temporal directions using bidirectional state-space models (SSMs). For each direction $m \in \{\text{forward}, \text{backward}\}$, the output is computed as:
\begin{equation}
y_m = \text{SSM}(X'_m), \quad y_m \in \mathbb{R}^{B \times N \times D'}.
\end{equation}

Each $y_m$ is then passed through a learnable gating mechanism, producing gated outputs $y'_{\text{forward}}$ and $y'_{\text{backward}}$. The final output of the $\ell$-th layer is computed by aggregating the two directions and applying a residual connection:
\begin{equation}
T_{\ell} = \text{Linear}_T \left( y'_{\text{forward}} + y'_{\text{backward}} \right) + T_{\ell-1}.
\end{equation}

Since the SSM~\cite{ssm} directly operates on token embeddings and captures global contextual dependencies, its output inherently encodes the relative importance of each token. Furthermore, Mamba's architecture, characterized by high-dimensional channel representations, facilitates fine-grained attention modeling across channels. In contrast to Transformers, which assign a single attention weight per head, Mamba implicitly distributes attention via its channel-wise activations, allowing the model to capture more nuanced semantic relationships among tokens.

\begin{figure}[t]
    \centering
    \includegraphics[width=1\linewidth]{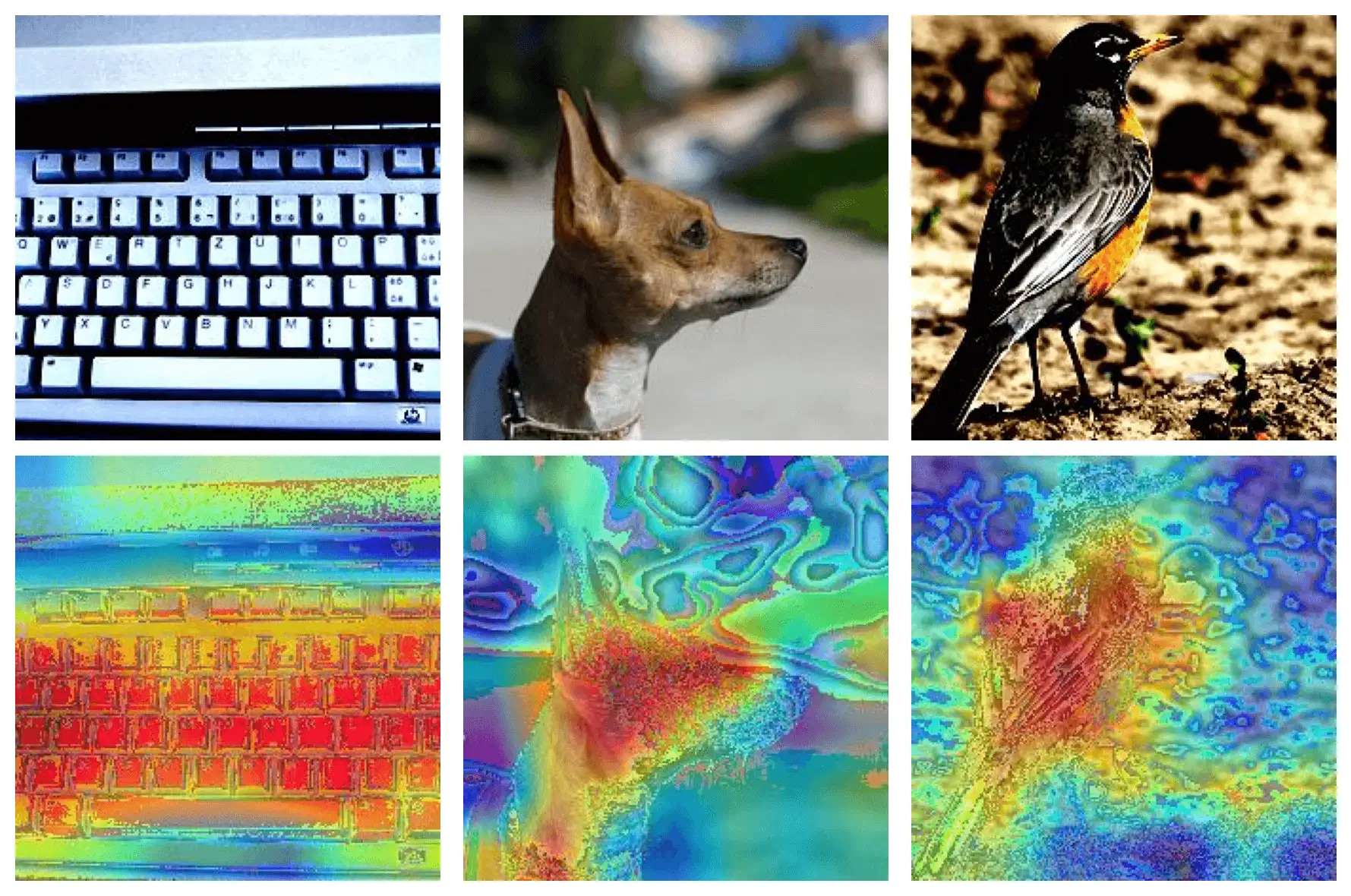}
    \caption{Visualization of the token-level attention map based on the proposed SSM-derived importance scores. Brighter regions indicate higher token importance, revealing the model's ability to focus on semantically salient areas.}
    \label{fig:score_visualization}
\end{figure}

Extending the previous exploration of the token importance metric~\cite{vimpruning1},
we further refine the formulation by introducing a Softplus-activated max-norm aggregation over the channel dimension of the SSM activations. The importance score $S$ for each token is thus computed as:
\begin{equation}
S = \max_{1 \le d \le D'}\, \operatorname{Softplus}\!\left(y_m^{\,d}\right),
\label{eq:token_score}
\end{equation}
where $\operatorname{Softplus}(x) = \ln(1 + e^{x})$ ensures non-negativity and smooth gradients, 
and $y_m^{\,d}$ denotes the $d$-th channel activation from the bidirectional SSM output. The effectiveness of this scoring mechanism is demonstrated in Figure~\ref{fig:score_visualization}, which presents an attention heatmap derived from the computed token importance scores for a representative input image.

Then informative regions can be identified using the token importance scores $S$ discussed above. 
To improve the stability of token importance estimation, we accumulate token importance scores across layers using an exponential moving average (EMA):
\begin{equation}
\bar{a}_n = \beta \cdot \bar{a}_{n-1} + (1 - \beta) \cdot a_0^n,
\label{eq:ema}
\end{equation}
where $\beta$ is the smoothing factor and $a_0^n$ denotes the importance score map in the $n$-th layer.

This aggregation begins from the final (12th) layer and includes every second Mamba layer in our design. The final set of informative tokens is selected based on the aggregated importance score vector $\bar{a}_N$ obtained from the last participating layer. 

Specifically, we rank all coarse patches based on their importance score values and select the top $\alpha N_c$ patches for refinement, where $\alpha \in [0, 1]$ is a hyperparameter that controls the refinement ratio. The total number of patches after this operation is computed as:
\begin{equation}
N_f = 4 \cdot \lceil \alpha N_c \rceil + \lfloor (1 - \alpha) N_c \rfloor,
\label{eq:patch_count}
\end{equation}

where $\lceil \cdot \rceil$ and $\lfloor \cdot \rfloor$ denote the ceiling and floor functions, respectively.

The parameter $\alpha$ governs the balance between computational cost and model accuracy. Setting $\alpha = 0$ disables fine-grained processing entirely, achieving maximum efficiency but potentially reducing recognition performance. In contrast, setting $\alpha = 1$ leads to full refinement of all coarse patches, effectively reducing the model to a standard ViM and incurring substantial computation. In our experiments, we choose $\alpha = 0.8$ to achieve a strong trade-off between accuracy and efficiency.


\subsubsection{Fine Region Feature Reuse} 
To further improve performance with minimal additional computation, we reuse the coarse-stage features corresponding to the selected fine regions. These features are fused to the fine-stage token sequence before being fed into the model for prediction.

In the fine stage, the input token sequence at the fine-grained stage is defined as:
\begin{equation}
\tilde{\mathcal{S}}_0 = [\tilde{s}_0^{(0)};\, \tilde{s}_1^{(0)};\, \dots;\, \tilde{s}_{N_f}^{(0)}] + \tilde{\mathcal{P}}_0,
\end{equation}
where each \(\tilde{s}_i^{(0)}\) is a fine-level patch token.

To enable effective fusion of coarse and fine-grained representations, it is essential to maintain spatial alignment between tokens across both stages. 
We first determine the spatial positions of coarse tokens selected for refinement. Each such position is then deterministically mapped to four adjacent fine-level tokens via a spatial index mapping function $\mathcal{I}: \{1, \ldots, N_c\} \rightarrow \{1, \ldots, N_f\}$. This guarantees that the resulting fine patches inherit the spatial structure of their corresponding coarse patch in the token sequence.

Given a flattened index $i \in \{0, 1, \ldots, N_c - 1\}$ corresponding to a selected coarse patch, we aim to compute the indices of its associated $2 \times 2$ fine-grained patches in the flattened fine grid.

Let $H_1 = \frac{\text{image\_size}}{\text{patch\_size}_1}$ and $H_2 = \frac{\text{image\_size}}{\text{patch\_size}_2}$ denote the spatial widths of the coarse and fine grids, respectively.

We first compute the column index $y$ of the coarse patch in the 2D grid:
\begin{equation}
y = i \bmod H_1.
\end{equation}

Then, the corresponding four fine-level patch indices in the flattened sequence are computed as:
\begin{align}
\mathcal{I}_1(i) = 4i - 2y,
\mathcal{I}_2(i) = \mathcal{I}_1(i) + 1, \nonumber \\
\mathcal{I}_3(i) = \mathcal{I}_1(i) + H_2,
\mathcal{I}_4(i) = \mathcal{I}_3(i) + 1.
\end{align}

This mapping ensures that each coarse patch is spatially aligned with its corresponding $2 \times 2$ fine patches in the flattened fine-grid sequence.


To maintain the spatial structure of the input, the selected coarse token indices are first sorted based on their original spatial order prior to mapping them to the fine-level sequence. This unified alignment strategy ensures coherent multi-scale feature integration, enhancing the model’s capacity to capture both global contextual information and fine-grained local details.

Specifically, each token from the coarse encoder outputs \([s_1^{(d)}, s_2^{(d)}, \dots, s_m^{(d)}]\) is passed through an MLP:
\begin{equation}
s_i^{\text{reuse}} = \text{MLP}(s_i^{(d)}),
\end{equation}
and is then reshaped and replicated four times to match the layout of the corresponding \(2 \times 2\) fine-level patches.

Tokens originating from unselected coarse patches, as well as the [CLS] token from the coarse stage, are zeroed out. The resulting reused feature sequence is expressed as:
\begin{equation}
\mathcal{S}_r = \text{FR}([s_1^{(d)};\, s_2^{(d)};\, \dots;\, s_m^{(d)}]),
\end{equation}
where \(\text{FR}(\cdot)\) represents the fine region feature reuse transformation combined with selective masking.

This reused feature is fused with the fine-stage input via a residual shortcut:
\begin{equation}
\Phi_f(\tilde{\mathcal{S}}_0 + \mathcal{S}_r) = [\tilde{s}_0^{(d)};\, \tilde{s}_1^{(d)};\, \dots;\, \tilde{s}_{N_f}^{(d)}],
\end{equation}
where \(\Phi_f\) is the fine-stage Vision Mamba encoder.

Finally, the fine-stage class token \(\tilde{s}_0^{(d)}\) is fed into the classifier \(F\) to produce the final prediction:
\begin{equation}
\mathbf{p} = F(\tilde{s}_0^{(d)}) = [p_1, p_2, \dots, p_C].
\end{equation}

\subsection{Training Procedure}
In training MambaScope, we fix the confidence threshold \(\eta\) to 1, such that all input samples are routed through the fine-grained inference stage. This design encourages the fine-level predictions to serve as strong supervision signals, ensuring accurate alignment with the ground-truth labels. At the same time, the model implicitly learns to make reliable coarse-stage predictions by minimizing their divergence from the fine-stage outputs. This helps the coarse branch approximate the behavior of the fine branch, which is essential for achieving efficient inference later.

The overall training objective combines both prediction accuracy and consistency between stages, and is defined as:
\begin{equation}
\mathcal{L} = \text{CE}(\mathbf{p}, y) + \text{KL}(\mathbf{q} \parallel \mathbf{p}),
\end{equation}
where \(\text{CE}(\cdot, \cdot)\) denotes the cross-entropy loss and \(\text{KL}(\cdot \parallel \cdot)\) represents the Kullback–Leibler divergence between the coarse  and the fine prediction.

During inference, the threshold \(\eta\) is treated as a tunable parameter that controls the trade-off between computational cost and prediction accuracy. Higher values of \(\eta\) lead to more samples undergoing fine-level processing, which improves performance but incurs greater computational expense. In contrast, lower thresholds allow more predictions to be made at the coarse stage, reducing computation with potentially minor drops in accuracy.

\section{Experiments}

\begin{table}[!t]
  \centering
  \small
  \caption{Comparison between existing token optimization strategies in ViM for image classification on ImageNet-1K.}
  \setlength{\tabcolsep}{0.2mm}{
  \begin{tabular}{@{}lccc@{}}
    \hline
    \textbf{Model} & Params & FLOPs (G) $\downarrow$ & Top-1 Acc. (\%) \\
        \hline
        \rowcolor[gray]{0.9}
    ViM-T & 7M & 1.5 & 76.1 \\
    ViM-T + HA & 7M & 1.3 (↓13.3\%) & 75.1 (-1.0) \\
    R-MeeTo-T & 7M & 1.3 (↓13.3\%) & 75.3 (-0.8) \\
    ViM-T + DyVM & 7M & 1.3 (↓13.3\%) & 75.2 (-0.9) \\
    \hdashline 
    \textbf{MambaScope-T (default)} & 8M & 1.4 (↓6.7\%) & \textbf{76.4} (+0.3)\\
    \hline
     \rowcolor[gray]{0.9}
    ViM-S & 26M & 5.1 & 80.3 \\
    ViM-S + HA & 27M & 3.6 (↓29.4\%) & 78.8 (-1.5) \\
    R-MeeTo-S & 26M & 3.6 (↓29.4\%) & 79.9 (-0.4)\\
    ViM-S + DyVM & 27M & 3.5 (↓31.4\%) & 78.8 (-1.5) \\
    \hdashline 
    \textbf{MambaScope-S ($\eta$ = 0.5}) & 28M & \textbf{2.5} (↓51.0\%) & 79.8 (-0.5) \\
    \textbf{MambaScope-S ($\eta$ = 0.55)} & 28M & 2.7 (↓47.1\%) & 80.3 (+0.0) \\
    \textbf{MambaScope-S ($\eta$ = 0.65})& 28M & 3.1 (↓39.2\%) & 80.7 (+0.4)\\
   \textbf{MambaScope-S (default)} & 28M & 4.3 (↓15.7\%) & \textbf{81.2} (+0.9)\\
    \hline
     \rowcolor[gray]{0.9}
    ViM-B & 98M & 18.9 & 81.9 \\
    R-MeeTo-B & - & 13.2 (↓30.2\%) & 81.3 (-0.6) \\
    ViM-B + DyVM & 101M & 12.1 (↓36.0\%) & 80.0 (-1.9) \\
    \hdashline 
    \textbf{MambaScope-B ($\eta$ = 0.7)} & 102M & \textbf{11.8} (↓37.6\%) & \textbf{82.8} (+0.9)\\
    \hline
  \end{tabular}
  
  \label{tab:maintable}
  }
\end{table}

  

\subsection{ImageNet-1K Classification}
\noindent\textbf{Settings.}
Our primary evaluation is conducted on the standard ImageNet~\cite{deng2009imagenet} benchmark, which contains approximately 1.28 million training images and 50,000 validation images across 1,000 categories. To ensure a fair comparison, we implement MambaScope on top of the ViM~\cite{vim} architecture, following prior work on ViM token pruning~\cite{wu2025dvmamba, vimpruning1, vimpruning2}. All training hyperparameters, including image preprocessing, learning rate, and optimizer settings, follow the standard configuration of ViM~\cite{vim} to ensure consistency. Notably, the coarse and fine stages share all network parameters. To bridge the resolution gap, coarse patches are downsampled to match the fine-stage embedding. For stable training, full-image fine splitting is used for the first 200 epochs, then limited to informative coarse patches in the remaining training epochs.

\begin{figure}[!t]
    \centering
    \includegraphics[width=0.48\textwidth]{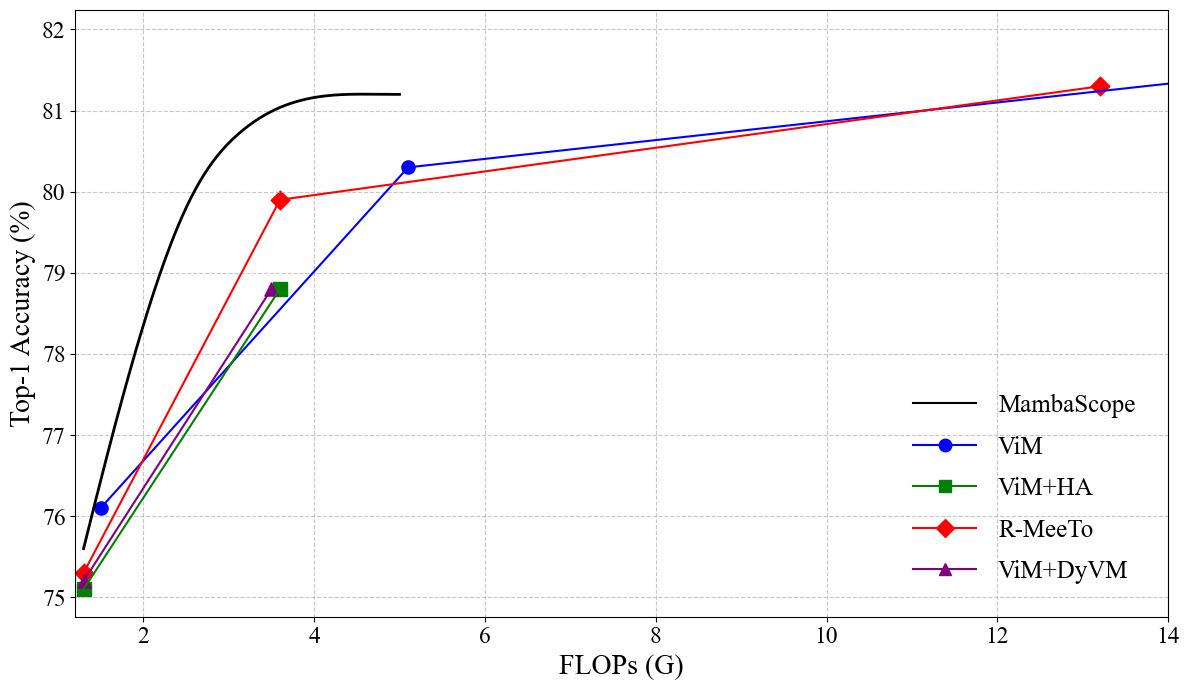}
    \caption{Comparison of Top-1 accuracy versus FLOPs among other ViM optimization methods.}
    \label{fig:compare}
\end{figure}

\noindent\textbf{Results.}
As shown in Table~\ref{tab:maintable} and Fig.~\ref{fig:compare}, MambaScope consistently surpasses prior token optimization methods based on the ViM backbone on ImageNet-1K. MambaScope-T achieves 76.4\% Top-1 accuracy with only 1.4G FLOPs, exceeding ViM-T (76.1\%) and lightweight variants such as DyVM (75.2\%) and HA (75.1\%) under similar budgets. MambaScope-S provides flexible efficiency–accuracy control via the confidence threshold~$\eta$: at $\eta$ = 0.5, 0.55, and 0.65, it attains 79.8\%, 80.3\%, and 80.7\% accuracy, while reducing FLOPs by 51.0\%, 47.1\%, and 39.2\% compared to ViM-S. At the default setting, MambaScope-S reaches 81.2\% with 15.7\% fewer FLOPs (4.3G vs. 5.1G), outperforming DyVM and HA (both 78.8\%). MambaScope-B further scales to 82.8\% accuracy with 37.6\% fewer FLOPs than ViM-B. Overall, MambaScope delivers strong scalability and superior efficiency–accuracy trade-offs, confirming the effectiveness of its coarse-to-fine token selection and adaptive refinement.



\begin{table}[t]
  \centering
  \small
  \setlength{\tabcolsep}{2.2mm}
  \caption{Semantic segmentation performance on the ADE20K \textit{val} set using the UperNet framework~\cite{upernet}.}
  \label{tab:ade20k}

  \begin{tabular}{l c c c c}
    \hline
    \textbf{Model} & Image Size & Params & FPS $\uparrow$ & mIoU \\
    \hline
    DeiT\mbox{-}Ti       & $512^{2}$ & 11M & 2.52 & 39.2 \\
    \rowcolor[gray]{0.9}
    Vim\mbox{-}Ti        & $512^{2}$ & 13M & 2.52 & 41.0 \\
    Vim\mbox{-}Ti+DyVM   & $512^{2}$ & --  & --   & 40.1 \\
    \textbf{MambaScope\mbox{-}Ti}      & $512^{2}$ & 13M & \textbf{2.91} & \textbf{41.3} \\
    \hline
    ResNet\mbox{-}50     & $512^{2}$ & 67M & --   & 41.2 \\
    DeiT\mbox{-}S        & $512^{2}$ & 43M & --   & 44.0 \\
    \rowcolor[gray]{0.9}
    Vim\mbox{-}S         & $512^{2}$ & 46M & --   & 44.9 \\
    Vim\mbox{-}S+DyVM    & $512^{2}$ & --  & --   & 42.0 \\
    \textbf{MambaScope\mbox{-}S}       & $512^{2}$ & 48M & --   & \textbf{45.5} \\
    \hline
  \end{tabular}
\end{table}

\subsection{Semantic Segmentation}

\noindent\textbf{Settings.}  
For semantic segmentation evaluation, we adopt the ADE20K dataset~\cite{ade20k} and employ UperNet~\cite{upernet} as the segmentation framework.  
The training is conducted for 160K iterations with a batch size of 16, using the AdamW optimizer with a learning rate of $6 \times 10^{-5}$.  
All experiments are performed at a default input resolution of $512 \times 512$.  
To further compare the efficiency of MambaScope with ViM on downstream tasks such as segmentation, detection, and instance segmentation, we integrate the backbones with a standard Feature Pyramid Network (FPN) module and benchmark their throughput in terms of FPS.

\noindent\textbf{Results.}  
As shown in Table~\ref{tab:ade20k}, MambaScope consistently outperforms prior backbones on ADE20K.  
MambaScope-Ti achieves 41.3~mIoU and 2.91~log-FPS, surpassing DeiT-Ti (39.2~mIoU, 2.52~log-FPS) and Vim-Ti (41.0~mIoU, 2.52~log-FPS) under similar parameter budgets.  
MambaScope-S attains the highest segmentation accuracy of 45.5~mIoU with comparable parameters to Vim-S (46M vs.~48M), validating the effectiveness of the coarse-to-fine design.
Overall, MambaScope delivers superior segmentation quality while maintaining high computational efficiency even with reduced token usage.

\subsection{Object Detection and Instance Segmentation}
\noindent\textbf{Settings.}  
We evaluate MambaScope on the COCO 2017 benchmark~\cite{coco2017} for both object detection and instance segmentation, adopting Cascade Mask R-CNN~\cite{casrcnn} as the detection framework.  
During training, we use the AdamW optimizer with a weight decay of 0.1 and a total batch size of 64. The training schedule follows 380K iterations with an initial learning rate of $1 \times 10^{-4}$ and linear decay.  
All backbones are initialized with weights pretrained on ImageNet-1K.

\noindent\textbf{Results.}
 As shown in Table~\ref{tab:coco}, MambaScope-Ti achieves strong performance across both tasks, obtaining 45.8 box AP and 39.4 mask AP—surpassing DeiT-Ti by +1.4 and +1.3, respectively—while maintaining accuracy comparable to ViM-Ti. In addition, MambaScope-Ti provides balanced gains across different object scales, particularly for medium and large objects. Compared with ViM-Ti, it reduces backbone computation by approximately 30G FLOPs without sacrificing accuracy. These results confirm that the proposed coarse-to-fine token reduction enhances spatial representation and long-range dependency modeling, leading to a more favorable efficiency–accuracy trade-off in dense prediction tasks.

\begin{table}[t]
  \centering
  \small
  \setlength{\tabcolsep}{0.1pt}
  \caption{Results of object detection and instance segmentation on the COCO \textit{val} set using Cascade Mask R-CNN~\cite{casrcnn}.}
  \begin{tabular}{l c c c c c c c c}
    \hline
    \textbf{Model} & Params & FLOPs $\downarrow$ & AP$^{\text{box}}$ & AP$^{\text{box}}_{50}$ & AP$^{\text{box}}_{75}$ & 
              AP$^{\text{box}}_{s}$ & AP$^{\text{box}}_{m}$ & AP$^{\text{box}}_{l}$ \\
    \hline
    DeiT-Ti     & -    & -    & 44.4 & 63.0 & 47.8 & 26.1 & 47.4 & 61.8 \\
    \rowcolor[gray]{0.9}
    Vim-Ti      & 30M    & 352G    & 45.7 & 63.9 & 49.6 & 26.1 & 49.0 & 63.2 \\
    \textbf{Ours-Ti}    & 30M  & \textbf{321G} & \textbf{45.8} & \textbf{64.5} & \textbf{49.6} & \textbf{26.2} & \textbf{49.2} & 63.1 \\
    \hline
    \textbf{Model} & Params & FLOPs $\downarrow$ & AP$^{\text{mask}}$ & AP$^{\text{mask}}_{50}$ & AP$^{\text{mask}}_{75}$ &
               AP$^{\text{mask}}_{s}$ & AP$^{\text{mask}}_{m}$ & AP$^{\text{mask}}_{l}$ \\
    \hline
    DeiT-Ti     & -    & -    & 38.1 & 59.9 & 40.5 & 18.1 & 40.5 & 58.4 \\
    \rowcolor[gray]{0.9}
    Vim-Ti      & 30M    & 352G    & 39.2 & 60.9 & 41.7 & 18.2 & 41.8 & 60.2 \\
    \textbf{Ours-Ti}    & 30M  & \textbf{321G} & \textbf{39.4} & \textbf{61.5} & \textbf{41.9} & \textbf{18.3} & \textbf{42.4} & \textbf{60.2} \\
    \hline
  \end{tabular}
  
  \label{tab:coco}
\end{table}

\subsection{Ablation Study}

\begin{table}[t]
  \centering
  \small
  \setlength{\tabcolsep}{1.1pt}
  \caption{Ablation study on different token importance metrics for MambaScope-S on the \textit{mini}ImageNet~\cite{miniimgnet} dataset.}
  \begin{tabular}{c|cccccc}
    \hline
    Metric & $\ell_{2}$-norm & w/o Clip & Clip & Softplus & Topk & Max-norm \\
    \hline
    Top-1 Acc. (\%) & 86.7 & 86.7 & 87.9 & 87.8 & 87.8 & \textbf{88.0} \\
    \hline
  \end{tabular}
  \label{tab:ablation-token-metric}
\end{table}

\begin{table}[t]
\small

\begin{minipage}{0.16\textwidth}
\centering
\setlength{\tabcolsep}{1.5mm} 
\captionof{table}{Effect of various refinement ratio $\alpha$.}
\label{tab:alpha_ablation}
\begin{tabular}{c c c}
\hline
$\alpha$ & Acc. & FLOPs \\
\hline
0.5 & 78.5 & 3.2 \\
0.6 & 79.1 & 3.5 \\
0.7 & 79.7 & 3.9 \\
\rowcolor[gray]{0.9}
0.8 & 80.1 & 4.4 \\
0.9 & 80.3 & 4.7 \\
\hline
\end{tabular}
\end{minipage}%
\hfill 
\begin{minipage}{0.15\textwidth}
\centering
\setlength{\tabcolsep}{1.5mm}
\captionof{table}{Effect of various aggregation weight $\beta$.}
\label{tab:beta_ablation}
\begin{tabular}{c c}
\hline
$\beta$ & Acc. \\
\hline
0     & 86.8 \\
0.5   & 87.1 \\
0.9   & 87.4 \\
0.99  & 87.6 \\
\rowcolor[gray]{0.9}
0.999 & 87.6 \\
\hline
\end{tabular}
\end{minipage}%
\hfill
\begin{minipage}{0.15\textwidth}
\centering
\setlength{\tabcolsep}{2mm}
\captionof{table}{Effect of various layer selection.}
\label{tab:layers_ablation}
\begin{tabular}{c c}
\hline
\#Layers & Acc. \\
\hline
6  & 86.7 \\
\rowcolor[gray]{0.9}
12 & 87.0 \\
18 & 86.8 \\
\hline
\end{tabular}
\end{minipage}

\end{table}




\textbf{Importance Metric.} 
Table~\ref{tab:ablation-token-metric} shows the evaluation of different token importance metrics on ViM-S using the \textit{mini}ImageNet~\cite{miniimgnet} dataset. 
The baseline $\ell_{2}$-norm and \textit{w/o Clip} achieve 86.7\% accuracy, while applying \textit{Clip}, \textit{Softplus}, or \textit{Top-k} yields moderate improvements (approximately 87.8--87.9\%). 
Among all methods, \textit{Max-norm} performs best with 88.0\% accuracy, demonstrating its superior ability to capture informative tokens and enhance pruning robustness.


\noindent\textbf{Refinement Ratio.} 
Table~\ref{tab:alpha_ablation} shows that increasing the refinement ratio $\alpha$ improves accuracy but raises FLOPs. For example, accuracy grows from 78.5\% to 80.3\% as $\alpha$ increases from 0.5 to 0.9, with FLOPs rising from 3.2G to 4.7G. We adopt $\alpha = 0.8$ as a balanced setting. 

\noindent\textbf{Aggregation Weight.} 
Table~\ref{tab:beta_ablation} shows that increasing the aggregation weight $\beta$, which incorporates more early-layer information, improves performance. Setting $\beta = 0.99$ achieves the best accuracy and is used as default.

\noindent\textbf{Layer Selection.} Table~\ref{tab:layers_ablation} evaluates how aggregation depth affects performance with $\alpha = 0.8$. Using the last 12 layers achieves the best accuracy (87.6\%), suggesting it balances semantic richness and score stability.

\begin{figure}[!t]
    \centering
    \includegraphics[width=0.48\textwidth]{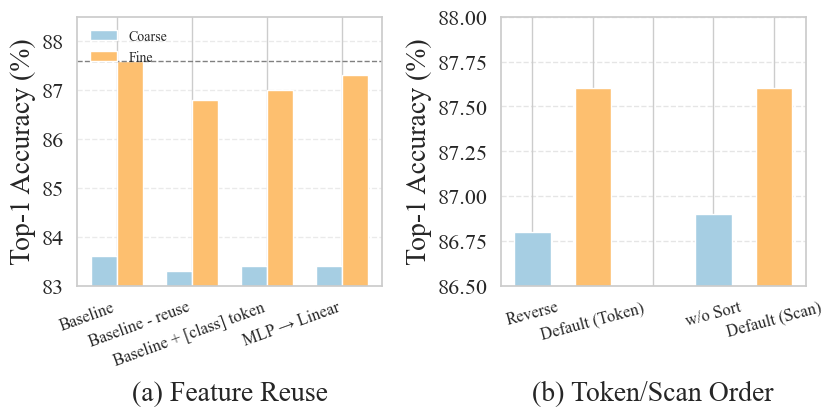}
    \caption{Ablation study on feature reuse and token arrangement strategies conducted on the \textit{mini}ImageNet~\cite{miniimgnet} dataset.}

    \label{fig:type2}
\end{figure}


\noindent\textbf{Feature Reuse.} Our method reuses coarse-stage visual tokens via a lightweight MLP while excluding the [class] token. As shown in Fig.~\ref{fig:type2}(a), reusing the [class] token or replacing the MLP with a linear layer reduces fine-stage accuracy to 87.0\% and 87.3\%, respectively, compared to 87.6\% with our design. These results underscore the value of selective reuse and non-linear transformation.


\noindent\textbf{Token Order.}  
Fig.~\ref{fig:type2}(b) compares placing the [CLS] token before or after the important tokens.  
Positioning [CLS] ahead of them (default) achieves higher accuracy (87.6\% vs. 86.8\%), indicating that early aggregation of key context improves recognition.



\noindent\textbf{Scanning Order.} Fig.~\ref{fig:type2}(b) compares token ordering strategies during the coarse-to-fine transition. Placing important tokens first yields higher accuracy (87.6\%) than the reverse (86.9\%), indicating that front-loading informative content enhances fine-stage recognition.

\begin{figure}[t]
    \centering
    \includegraphics[width=0.48\textwidth]{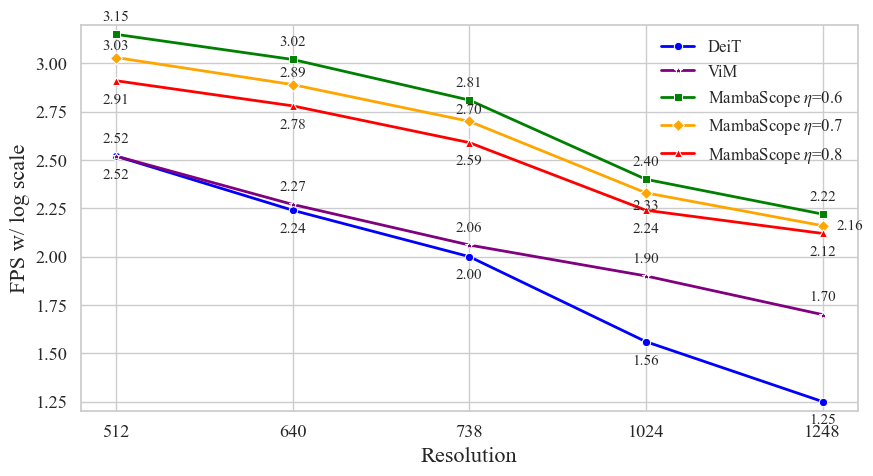}
    \caption{Log-scaled FPS comparison between baselines (DeiT/Vim) and MambaScope variants under different thresholds.}

    \label{fig:fps}
\end{figure}

\noindent\textbf{Confidence Threshold.} 
Figure~\ref{fig:fps} shows the FPS comparison with different thresholds $\eta \in \{0.6, 0.7, 0.8\}$ across multiple resolutions on the semantic segmentation task, where computational cost increases much more rapidly with input resolution than in image classification. 
Varying~$\eta$ leads to notable changes in runtime. Meanwhile, all MambaScope variants consistently outperform DeiT and ViM in inference speed, and this advantage becomes more evident as the resolution increases. These results demonstrate that the confidence threshold~$\eta$ enables flexible control of computational cost, while simultaneously verifying the strong resolution scalability and efficiency of our MambaScope.

\subsection{Visualization}
\begin{figure}[!t]
    \centering
    \includegraphics[width=0.48\textwidth]{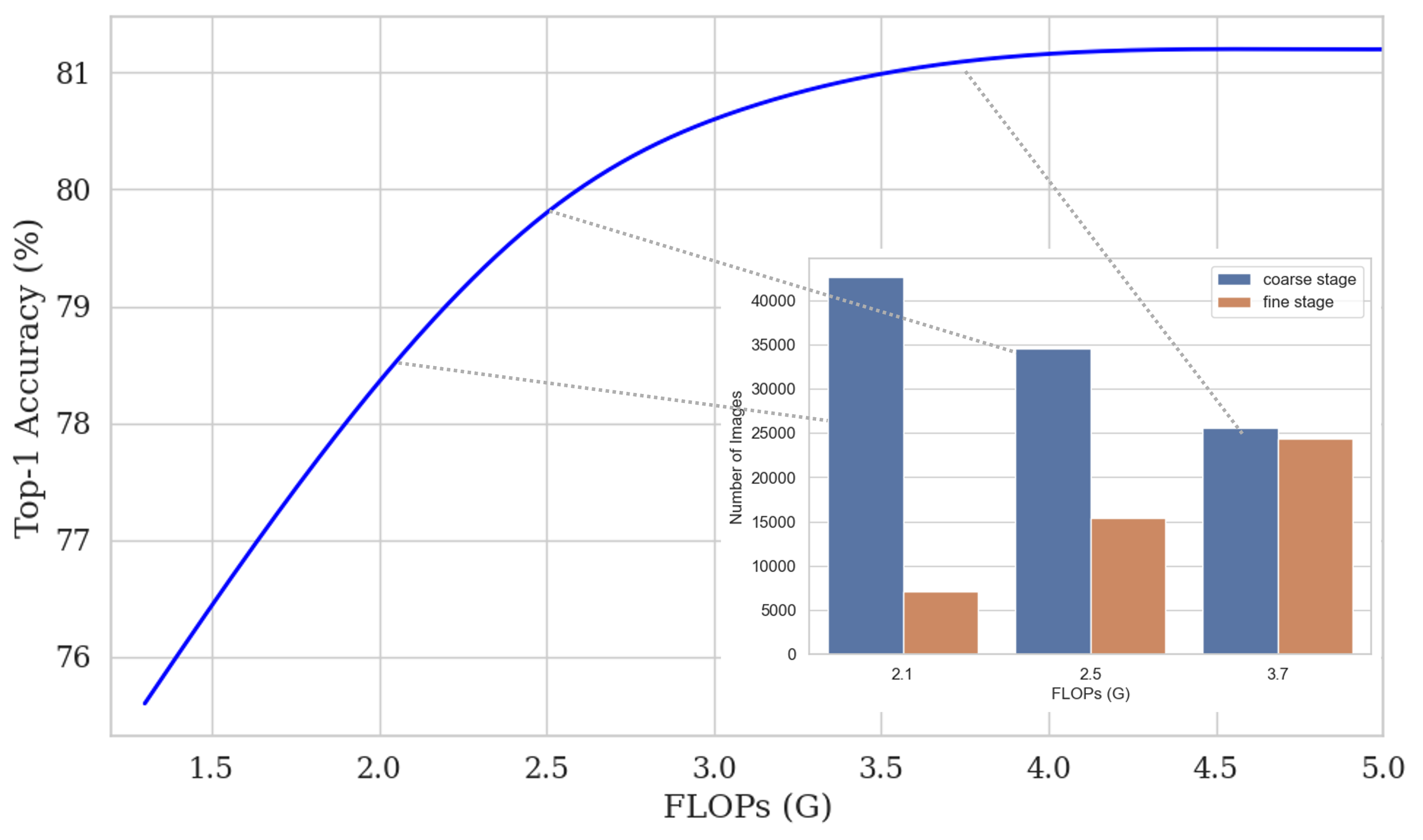}
    \caption{Accuracy trend and image distribution across coarse and fine stages.}
    \label{fig:comparetrend}
\end{figure}

Fig.~\ref{fig:cfcompare} shows representative examples correctly recognized by our model at both coarse and fine stages. The left column presents coarse-stage predictions, and the right displays refined fine-stage results, where only informative regions are visualized for clarity. Images confidently recognized at the coarse stage usually contain salient objects in simple backgrounds, while those routed to the fine stage involve cluttered scenes or small, ambiguous targets. In such cases, selective refinement further enhances token representations in critical regions, improving overall recognition. The refined areas align closely with target objects, confirming the effectiveness of our adaptive token selection strategy. Fig.~\ref{fig:comparetrend} further shows the distribution of correctly classified images at each stage. Adjusting the threshold~$\eta$ enables MambaScope to flexibly balance computation and accuracy, with higher~$\eta$ values routing more samples to fine refinement.

\begin{figure}[t]
    \centering
    \includegraphics[width=0.48\textwidth]{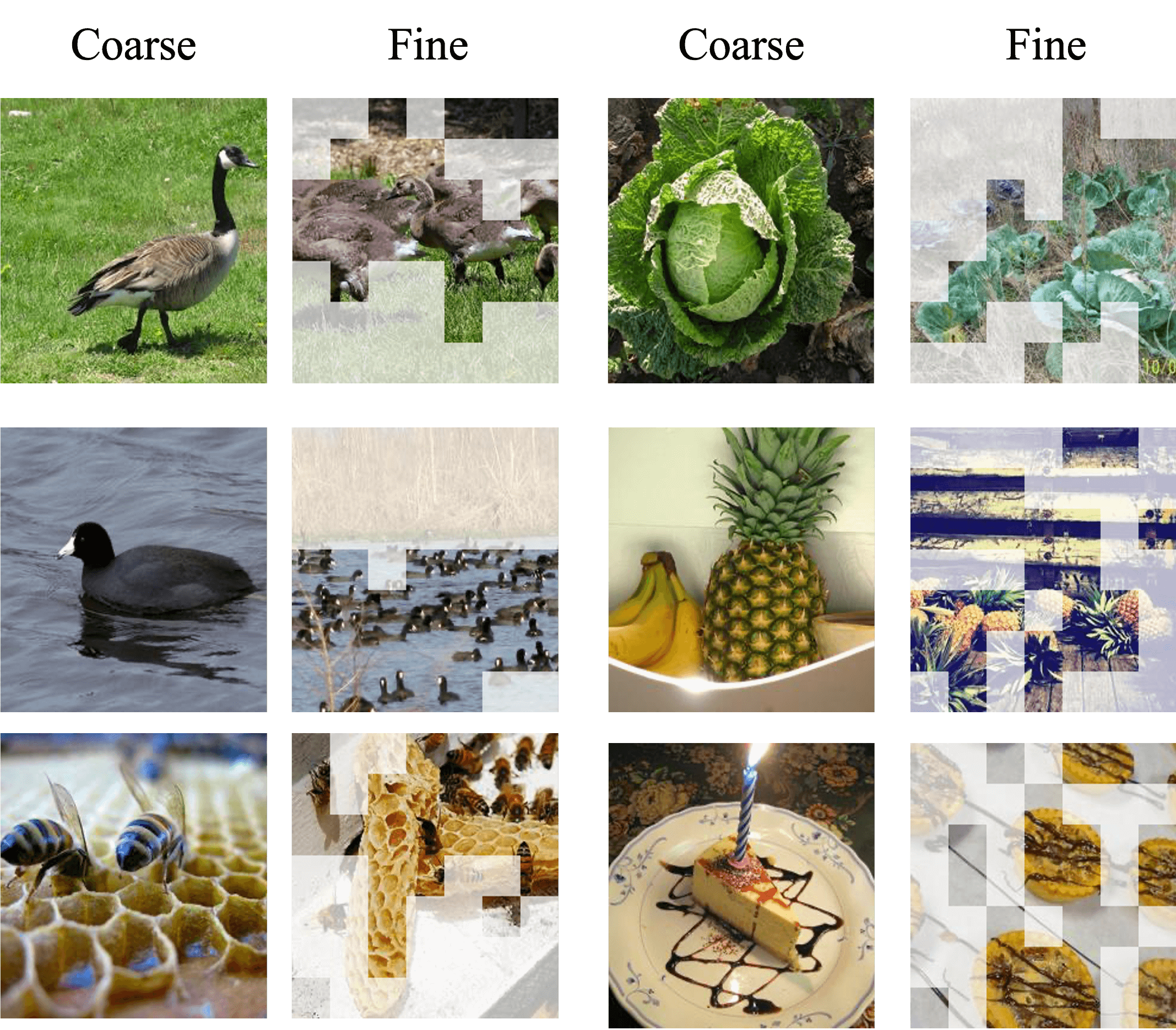}
    \caption{Visualization of images that just run through the coarse inference stage (Coarse) and both two inference stages (Fine).}
    \label{fig:cfcompare}
\end{figure}

\section{Conclusion}
We propose MambaScope, a coarse-to-fine scoping framework that improves the efficiency of Vision Mamba models through adaptive resolution inference. MambaScope performs coarse prediction for visually simple samples and selectively refines informative regions at higher resolution when model confidence is low. This adaptive mechanism reduces redundant token computation while preserving essential visual details, achieving efficiency gains without architectural changes or training–inference inconsistency. Extensive experiments on ImageNet, ADE20K, and COCO 2017 show that MambaScope delivers superior accuracy–efficiency trade-offs compared with ViM and recent token reduction baselines. Its feature-reuse and scan-order strategies further improve spatial consistency and generalize effectively across Mamba-based backbones and dense prediction tasks.

{
    \small
    \bibliographystyle{ieeenat_fullname}
    \bibliography{main}

@String(CVPR= {IEEE Conf. Comput. Vis. Pattern Recog.})

@String(ICCV= {Int. Conf. Comput. Vis.})

@String(ECCV= {Eur. Conf. Comput. Vis.})

@String(AAAI = {AAAI})

@String(CVPR  = {CVPR})

@String(ICCV  = {ICCV})

@String(ECCV  = {ECCV})

@misc{vimpruning1,
      title={Exploring Token Pruning in Vision State Space Models}, 
      author={Zheng Zhan and Zhenglun Kong and Yifan Gong and Yushu Wu and Zichong Meng and Hangyu Zheng and Xuan Shen and Stratis Ioannidis and Wei Niu and Pu Zhao and Yanzhi Wang},
      year={2024},
      eprint={2409.18962},
      archivePrefix={arXiv},
      primaryClass={cs.CV},
      url={https://arxiv.org/abs/2409.18962}, 
}

@misc{vimpruning2,
      title={Dynamic Vision Mamba}, 
      author={Mengxuan Wu and Zekai Li and Zhiyuan Liang and Moyang Li and Xuanlei Zhao and Samir Khaki and Zheng Zhu and Xiaojiang Peng and Konstantinos N. Plataniotis and Kai Wang and Wangbo Zhao and Yang You},
      year={2025},
      eprint={2504.04787},
      archivePrefix={arXiv},
      primaryClass={cs.CV},
      url={https://arxiv.org/abs/2504.04787}, 
}

@misc{vimmerge,
      title={Faster Vision Mamba is Rebuilt in Minutes via Merged Token Re-training}, 
      author={Mingjia Shi and Yuhao Zhou and Ruiji Yu and Zekai Li and Zhiyuan Liang and Xuanlei Zhao and Xiaojiang Peng and Shanmukha Ramakrishna Vedantam and Wangbo Zhao and Kai Wang and Yang You},
      year={2025},
      eprint={2412.12496},
      archivePrefix={arXiv},
      primaryClass={cs.CV},
      url={https://arxiv.org/abs/2412.12496}, 
}

@inproceedings{vim,
  title={Vision mamba: efficient visual representation learning with bidirectional state space model},
  author={Zhu, Lianghui and Liao, Bencheng and Zhang, Qian and Wang, Xinlong and Liu, Wenyu and Wang, Xinggang},
  booktitle={Proceedings of the 41st International Conference on Machine Learning},
  pages={62429--62442},
  year={2024}
}

@inproceedings{attn,
  title     = {Attention Is All You Need},
  author    = {Vaswani, Ashish and Shazeer, Noam and Parmar, Niki and Uszkoreit, Jakob and Jones, Llion and Gomez, Aidan N and Kaiser, Lukasz and Polosukhin, Illia},
  booktitle = {Advances in Neural Information Processing Systems (NeurIPS)},
  volume    = {30},
  year      = {2017},
  publisher = {Curran Associates, Inc.}
}

@article{vim4,
  title   = {EfficientVMamba: Atrous Selective Scan for Light Weight Visual Mamba},
  author  = {Pei, Xiaohuan and Huang, Tao and Xu, Chang},
  journal = {arXiv preprint arXiv:2403.09977},
  year    = {2024},
  url     = {https://arxiv.org/abs/2403.09977}
}

@article{vmamba,
  title   = {vMamba: Visual State Space Model},
  author  = {Liu, Yue and Tian, Yunjie and Zhao, Yuzhong and Yu, Hongtian and Xie, Lingxi and Wang, Yaowei and Ye, Qixiang and Liu, Yunfan},
  journal = {arXiv preprint arXiv:2401.10166},
  year    = {2024},
  url     = {https://arxiv.org/abs/2401.10166}
}

@article{plainmamba,
  title   = {PlainMamba: Improving Non-Hierarchical Mamba in Visual Recognition},
  author  = {Yang, Chenhongyi and Chen, Zehui and Espinosa, Miguel and Ericsson, Linus and Wang, Zhenyu and Liu, Jiaming and Crowley, Elliot J.},
  journal = {arXiv preprint arXiv:2403.17695},
  year    = {2024},
  url     = {https://arxiv.org/abs/2403.17695}
}

@article{huang2024localmamba,
  title   = {LocalMamba: Visual State Space Model with Windowed Selective Scan},
  author  = {Huang, Tao and Pei, Xiaohuan and You, Shan and Wang, Fei and Qian, Chen and Xu, Chang},
  journal = {arXiv preprint arXiv:2403.09338},
  year    = {2024},
  url     = {https://arxiv.org/abs/2403.09338}
}

@article{xie2024quadmamba,
  title   = {QuadMamba: Learning Quadtree-based Selective Scan for Visual State Space Model},
  author  = {Xie, Fei and Zhang, Weijia and Wang, Zhongdao and Ma, Chao},
  journal = {arXiv preprint arXiv:2410.06806},
  year    = {2024},
  url     = {https://arxiv.org/abs/2410.06806}
}

@article{mamba,
  title   = {Mamba: Linear-time Sequence Modeling with Selective State Spaces},
  author  = {Gu, Albert and Dao, Tri},
  journal = {arXiv preprint arXiv:2312.00752},
  year    = {2023},
  url     = {https://arxiv.org/abs/2312.00752}
}

@article{ssm,
  title   = {Efficiently Modeling Long Sequences with Structured State Spaces},
  author  = {Gu, Albert and Goel, Karan and R{\'e}, Christopher},
  journal = {arXiv preprint arXiv:2111.00396},
  year    = {2021},
  url     = {https://arxiv.org/abs/2111.00396}
}

@inproceedings{ssm2,
  title     = {Combining Recurrent, Convolutional, and Continuous-time Models with Linear State Space Layers},
  author    = {Gu, Albert and Johnson, Isys and Goel, Karan and Saab, Khaled and Dao, Tri and Rudra, Atri and R{\'e}, Christopher},
  booktitle = {Advances in Neural Information Processing Systems (NeurIPS)},
  volume    = {34},
  pages     = {572--585},
  year      = {2021}
}

@article{ssm4,
  title   = {On the Parameterization and Initialization of Diagonal State Space Models},
  author  = {Gu, Albert and Gupta, Ankit and Goel, Karan and R{\'e}, Christopher},
  journal = {arXiv preprint arXiv:2206.11893},
  year    = {2022},
  url     = {https://arxiv.org/abs/2206.11893}
}

@inproceedings{rao2021dynamicvit,
  title     = {DynamicViT: Efficient Vision Transformers with Dynamic Token Sparsification},
  author    = {Rao, Yulin and Zhao, Wenliang and Liu, Bing and Lu, Jie and Zhou, Jianmin and Hsieh, Cho-Jui},
  booktitle = {Advances in Neural Information Processing Systems (NeurIPS)},
  year      = {2021},
  volume    = {34}
}

@inproceedings{pan2021iared2,
  title     = {IA-RED\textsuperscript{2}: Interpretability-Aware Redundancy Reduction for Vision Transformers},
  author    = {Pan, Bohan and Panda, Rajeev and Jiang, Yanghao and Wang, Zhe and Feris, Rogerio and Oliva, Aude},
  booktitle = {Advances in Neural Information Processing Systems (NeurIPS)},
  year      = {2021},
  volume    = {34}
}

@article{bolya2023tokmerging,
  title   = {Token Merging: Your ViT But Faster},
  author  = {Bolya, Daniel and Fu, Cheng-Yang and Dai, Xiaoliang and Zhang, Peizhao and Feichtenhofer, Christoph and Hoffman, Judy},
  journal = {arXiv preprint arXiv:2302.12066},
  year    = {2023},
  url     = {https://arxiv.org/abs/2302.12066}
}

@inproceedings{xu2022evovit,
  title     = {Evo-ViT: Slow-Fast Token Evolution for Dynamic Vision Transformer},
  author    = {Xu, Yinglin and Zhang, Zhizhong and Zhang, Minghang and Sheng, Kai and Li, Ke and Dong, Wei and Zhang, Lei and Xu, Chang and Sun, Xian},
  booktitle = {Proceedings of the AAAI Conference on Artificial Intelligence (AAAI)},
  year      = {2022}
}

@inproceedings{wang2021dynamic,
  title     = {Not All Images Are Worth 16x16 Words: Dynamic Transformers for Efficient Image Recognition},
  author    = {Wang, Yulin and Huang, Rui and Song, Shiji and Huang, Zhiwu and Huang, Gao},
  booktitle = {Advances in Neural Information Processing Systems (NeurIPS)},
  year      = {2021},
  volume    = {34}
}

@article{zhan2024tokenpruning,
  title   = {Exploring Token Pruning in Vision State Space Models},
  author  = {Zhan, Zheng and Kong, Zhenglun and Gong, Yifan and Wu, Yushu and Meng, Zichong and Zheng, Hangyu and Shen, Xuan and Ioannidis, Stratis and Niu, Wei and Zhao, Pu and Wang, Yanzhi},
  journal = {arXiv preprint},
  year    = {2024},
  note    = {Manuscript},
}

@article{wu2025dvmamba,
  title   = {Dynamic Vision Mamba},
  author  = {Wu, Mengxuan and Li, Zekai and Liang, Zhiyuan and Li, Moyang and Zhao, Xuanlei and Khaki, Samir and Zhu, Zheng and Peng, Xiaojiang and Plataniotis, Konstantinos N. and Wang, Kai and others},
  journal = {arXiv preprint arXiv:2504.04787},
  year    = {2025},
  url     = {https://arxiv.org/abs/2504.04787}
}

@article{shi2024fastmamba,
  title   = {Faster Vision Mamba is Rebuilt in Minutes via Merged Token Re-training},
  author  = {Shi, Meng and Zhou, Yiyang and Yu, Rui and Li, Zekai and Liang, Zhiyuan and Zhao, Xuanlei and Peng, Xiaojiang and Vedantam, Satyanarayan R. and Zhao, Wenjun and Wang, Kai and others},
  journal = {arXiv preprint arXiv:2412.12496},
  year    = {2024},
  url     = {https://arxiv.org/abs/2412.12496}
}

@article{othermamba1,
  title   = {MambaVision: A Hybrid Mamba-Transformer Vision Backbone},
  author  = {Hatamizadeh, Ali and Kautz, Jan},
  journal = {arXiv preprint arXiv:2407.08083},
  year    = {2024},
  url     = {https://arxiv.org/abs/2407.08083}
}

@article{othermamba3,
  title   = {MambaND: Selective State Space Modeling for Multi-Dimensional Data},
  author  = {Li, Shufan and Singh, Harkanwar and Grover, Aditya},
  journal = {arXiv preprint arXiv:2402.05892},
  year    = {2024},
  url     = {https://arxiv.org/abs/2402.05892}
}

@inproceedings{deng2009imagenet,
  title     = {ImageNet: A Large-Scale Hierarchical Image Database},
  author    = {Deng, Jia and Dong, Wei and Socher, Richard and Li, Li-Jia and Li, Kai and Fei-Fei, Li},
  booktitle = {2009 IEEE Conference on Computer Vision and Pattern Recognition (CVPR)},
  pages     = {248--255},
  year      = {2009},
  organization = {IEEE}
}

@article{dosovitskiy2020vit,
  title   = {An Image is Worth 16x16 Words: Transformers for Image Recognition at Scale},
  author  = {Dosovitskiy, Alexey and Beyer, Lucas and Kolesnikov, Alexander and Weissenborn, Dirk and Zhai, Xiaohua and Unterthiner, Thomas and Dehghani, Mostafa and Minderer, Matthias and Heigold, Georg and Gelly, Sylvain and Uszkoreit, Jakob and Houlsby, Neil},
  journal = {arXiv preprint arXiv:2010.11929},
  year    = {2020},
  url     = {https://arxiv.org/abs/2010.11929}
}

@article{han2021dynamicnn,
  title   = {Dynamic Neural Networks: A Survey},
  author  = {Han, Yizeng and Huang, Gao and Song, Shiji and Yang, Le and Wang, Honghui and Wang, Yulin},
  journal = {IEEE Transactions on Pattern Analysis and Machine Intelligence (TPAMI)},
  volume  = {44},
  pages   = {7436--7456},
  year    = {2021},
  doi     = {10.1109/TPAMI.2021.3115310}
}

@article{liang2022evit,
  title   = {Not All Patches Are What You Need: Expediting Vision Transformers via Token Reorganizations},
  author  = {Liang, Youwei and Ge, Chongjian and Tong, Zhan and Song, Yibing and Wang, Jue and Xie, Pengtao},
  journal = {arXiv preprint arXiv:2202.07800},
  year    = {2022},
  url     = {https://arxiv.org/abs/2202.07800}
}

@inproceedings{meng2021adavit,
  title     = {AdaViT: Adaptive Vision Transformers for Efficient Image Recognition},
  author    = {Meng, Lingchen and Li, Hengduo and Chen, Bor-Chun and Lan, Shiyi and Wu, Zuxuan and Jiang, Yu-Gang and Lim, Ser Nam},
  booktitle = {Proceedings of the IEEE/CVF Conference on Computer Vision and Pattern Recognition (CVPR)},
  pages     = {12299--12308},
  year      = {2021}
}

@inproceedings{song2023dge,
  title     = {Dynamic Grained Encoder for Vision Transformers},
  author    = {Song, Lin and Zhang, Songyang and Liu, Songtao and Li, Zeming and He, Xuming and Sun, Hongbin and Sun, Jian and Zheng, Nanning},
  booktitle = {Advances in Neural Information Processing Systems (NeurIPS)},
  year      = {2023}
}

@inproceedings{liu2023revisiting,
  title     = {Revisiting Token Pruning for Object Detection and Instance Segmentation},
  author    = {Liu, Yifei and Gehrig, Mathias and Messikommer, Nico and Cannici, Marco and Scaramuzza, Davide},
  booktitle = {Proceedings of the IEEE/CVF Winter Conference on Applications of Computer Vision (WACV)},
  pages     = {2646--2656},
  year      = {2023}
}

@inproceedings{kong2022spvit,
  title     = {SPViT: Enabling Faster Vision Transformers via Soft Token Pruning},
  author    = {Kong, Zhenglun and Dong, Peiyan and Ma, Xiaolong and Meng, Xin and Niu, Wei and Sun, Mengshu and Ren, Bin and Qin, Minghai and Tang, Hao and Wang, Yanzhi},
  booktitle = {Proceedings of the European Conference on Computer Vision (ECCV)},
  year      = {2022}
}

@inproceedings{tvt,
  title     = {Tokens-to-Token ViT: Training Vision Transformers from Scratch on ImageNet},
  author    = {Yuan, Li and Chen, Yunpeng and Wang, Tao and Yu, Weihao and Shi, Yujun and Jiang, Zi-Hang and Tay, Francis EH and Feng, Jiashi and Yan, Shuicheng},
  booktitle = {Proceedings of the IEEE/CVF International Conference on Computer Vision (ICCV)},
  pages     = {558--567},
  year      = {2021}
}

@article{renggli2022mergetokens,
  title   = {Learning to Merge Tokens in Vision Transformers},
  author  = {Renggli, Cedric and Pinto, André Susano and Houlsby, Neil and Mustafa, Basil and Puigcerver, Joan and Riquelme, Carlos},
  journal = {arXiv preprint arXiv:2202.12015},
  year    = {2022},
  url     = {https://arxiv.org/abs/2202.12015}
}

@article{mehta2022gss,
  title   = {Long Range Language Modeling via Gated State Spaces},
  author  = {Mehta, Harsh and Gupta, Ankit and Cutkosky, Ashok and Neyshabur, Behnam},
  journal = {arXiv preprint arXiv:2206.13947},
  year    = {2022},
  url     = {https://arxiv.org/abs/2206.13947}
}

@article{smith2022s4d,
  title   = {Simplified State Space Layers for Sequence Modeling},
  author  = {Smith, Jimmy TH and Warrington, Andrew and Linderman, Scott W},
  journal = {arXiv preprint arXiv:2208.04933},
  year    = {2022},
  url     = {https://arxiv.org/abs/2208.04933}
}

@misc{upernet,
      title={Unified Perceptual Parsing for Scene Understanding}, 
      author={Tete Xiao and Yingcheng Liu and Bolei Zhou and Yuning Jiang and Jian Sun},
      year={2018},
      eprint={1807.10221},
      archivePrefix={arXiv},
      primaryClass={cs.CV},
      url={https://arxiv.org/abs/1807.10221}, 
}

@misc{miniimgnet,
      title={Matching Networks for One Shot Learning}, 
      author={Oriol Vinyals and Charles Blundell and Timothy Lillicrap and Koray Kavukcuoglu and Daan Wierstra},
      year={2017},
      eprint={1606.04080},
      archivePrefix={arXiv},
      primaryClass={cs.LG},
      url={https://arxiv.org/abs/1606.04080}, 
}

@misc{casrcnn,
      title={Cascade R-CNN: High Quality Object Detection and Instance Segmentation}, 
      author={Zhaowei Cai and Nuno Vasconcelos},
      year={2019},
      eprint={1906.09756},
      archivePrefix={arXiv},
      primaryClass={cs.CV},
      url={https://arxiv.org/abs/1906.09756}, 
}

@misc{coco2017,
      title={Microsoft COCO: Common Objects in Context}, 
      author={Tsung-Yi Lin and Michael Maire and Serge Belongie and Lubomir Bourdev and Ross Girshick and James Hays and Pietro Perona and Deva Ramanan and C. Lawrence Zitnick and Piotr Dollár},
      year={2015},
      eprint={1405.0312},
      archivePrefix={arXiv},
      primaryClass={cs.CV},
      url={https://arxiv.org/abs/1405.0312}, 
}

@misc{ade20k,
      title={Semantic Understanding of Scenes through the ADE20K Dataset}, 
      author={Bolei Zhou and Hang Zhao and Xavier Puig and Tete Xiao and Sanja Fidler and Adela Barriuso and Antonio Torralba},
      year={2018},
      eprint={1608.05442},
      archivePrefix={arXiv},
      primaryClass={cs.CV},
      url={https://arxiv.org/abs/1608.05442}, 
}

@misc{mambasurvy,
      title={Visual Mamba: A Survey and New Outlooks}, 
      author={Rui Xu and Shu Yang and Yihui Wang and Yu Cai and Bo Du and Hao Chen},
      year={2024},
      eprint={2404.18861},
      archivePrefix={arXiv},
      primaryClass={cs.CV}
}

@misc{cfvit,
      title={CF-ViT: A General Coarse-to-Fine Method for Vision Transformer}, 
      author={Mengzhao Chen and Mingbao Lin and Ke Li and Yunhang Shen and Yongjian Wu and Fei Chao and Rongrong Ji},
      year={2022},
      eprint={2203.03821},
      archivePrefix={arXiv},
      primaryClass={cs.CV},
      url={https://arxiv.org/abs/2203.03821}, 
}
}

\appendix
\maketitlesupplementary

\definecolor{mygreen}{RGB}{89,144,144}

\section{Implementation Details}
Our experiments span three major vision tasks: ImageNet-1K classification, ADE20K semantic segmentation, and COCO object detection and instance segmentation. For ImageNet-1K, we follow the standard ViM training configuration and optimize models on $224\times224$ inputs using AdamW (momentum 0.9, weight decay 0.05, batch size 256) with a cosine decay schedule from an initial learning rate of $1\times10^{-3}$ and EMA over 300 epochs. Training employs common augmentation strategies such as random cropping, horizontal flipping, Mixup, label smoothing, and random erasing. The first 200 epochs treat all patches as fine tokens, while the final 100 epochs activate informative patch selection using the hyperparameter~$\alpha$; at inference, we apply center cropping and evaluate under different routing thresholds~$\eta$. This setup ensures a consistent comparison with existing ViM-based token optimization methods and provides controlled evaluation of adaptive token usage.

For semantic segmentation on ADE20K, we adopt UperNet and train for 160K iterations using AdamW (weight decay 0.01, batch size 16) with 1500-step linear warmup and subsequent decay, using $512\times512$ inputs. To support early-exit inference, an auxiliary image-level classification head is incorporated during training, supervised with multi-label targets extracted from the segmentation masks. This auxiliary supervision enhances global semantic awareness without altering the standard segmentation framework. Data augmentation includes random flipping, scale jittering within [0.5, 2.0], and photometric distortions, and evaluation resizes images so that the shorter side is 512 pixels.

For COCO detection and instance segmentation, we use Cascade Mask R-CNN and train for 380K iterations using AdamW (weight decay 0.1, batch size 64). Large-scale jittering is applied to produce $1024\times1024$ inputs, and images are resized during evaluation to a 1024-pixel short side. These settings follow widely adopted detection protocols, ensuring that the improvements brought by our adaptive design are isolated from confounding factors related to training schedules or augmentation choices.

\section{More Visualized Results}

Figure~\ref{fig:cfcompare} presents additional visualizations illustrating the effectiveness of our method. While ViM extracts general visual features for classification, MambaScope selectively refines semantically important coarse tokens into fine-grained representations. This behavior enables the model to focus on critical areas, improving both efficiency and accuracy. The results validate the effectiveness of our coarse-to-fine token processing strategy.

\begin{algorithm}[t]
\small
\caption{Feature Reuse from Coarse to Fine Stage}
\label{alg:feature_reuse}
\begin{algorithmic}[1]
\State \textbf{Input:} coarse features $z_c$, fine features $x$
\State \textbf{Output:} fused features $x_f$
\State \textbf{Notation:} $B$ = batch size, $M$ = token length, $C$ = channel dimension

\State \textcolor{mygreen}{\textit{\# Fine-stage patch embedding and CLS insertion}}
\State $x = \text{patch\_embed}(x)$
\State $B, M, C = x.\text{shape}$
\State $cls\_token = \text{expand}(cls\_token, B, 1, C)$
\State $x = \text{concat}(x[:, :M//2, :], cls\_token, x[:, M//2:, :])$
\Statex
\State \textcolor{mygreen}{\textit{\# Transform coarse features and upsample}}
\State $z_c = \text{reuse\_block}(z_c)$
\State $z_c = \text{interpolate}(\text{reshape}(\text{transpose}(z_c, 1, 2), B, C, \sqrt{N}, \sqrt{N}), \text{size}=(patch\_h, patch\_w), \text{mode}=\texttt{nearest})$

\State $z_c = \text{transpose}(\text{reshape}(z_c, B, C, M), 1, 2)$
\Statex
\State \textcolor{mygreen}{\textit{\# Zero pad CLS position in reused features and fusion}}
\State $zero = \text{zeros}(B, 1, C)$
\State $z_c = \text{concat}(z_c[:, :M//2, :], zero, z_c[:, M//2:, :])$
\State $x_f = x + z_c$
\end{algorithmic}
\end{algorithm}

\begin{algorithm}[t]
\small
\caption{Informative Patch Selection}
\label{alg:informative_selection}
\begin{algorithmic}[1]
\State \textbf{Input:} Importance scores $A$, coarse tokens $z_c$, fine tokens $x$
\State \textbf{Output:} Selected and fused tokens $x$
\State \textbf{Notation:} $\alpha$: selection ratio, $N$: number of coarse tokens, $B$: batch size, $N'$: number of selected fine-grained tokens

\Statex
\State \textcolor{mygreen}{\textit{\# Sort tokens by importance (excluding CLS)}}
\State $A = \text{remove\_cls}(A)$
\State $k = \lceil \alpha \cdot N \rceil$
\State $policy = \text{argsort}(A, \text{desc=True})$
\State $imp, unimp = \text{sort}(policy[:, :k]), \text{sort}(policy[:, k:])$
\Statex
\State \textcolor{mygreen}{\textit{\# Extract unimportant coarse tokens}}
\State $z_{unimp} = \text{select}(z_c, \text{remap\_cls}(unimp))$
\Statex
\State \textcolor{mygreen}{\textit{\# Convert important coarse indices to fine indices}}
\State $imp\_fine = \text{sort}(\text{reshape}(\text{coarse2fine}(imp), B, -1))$
\State $x_{imp} = \text{select}(x, \text{remap\_cls}(imp\_fine))$
\Statex
\State \textcolor{mygreen}{\textit{\# Insert CLS token and fusion.}}
\State $cls = \text{expand}(cls\_token, B, 1, C)$
\State $m = N' // 2$
\State $x_{imp}^{cls} = \text{concat}(x_{imp}[:, :m, :], cls, x_{imp}[:, m:, :])$
\State $x = \text{concat}(x_{imp}^{cls}, z_{unimp})$
\end{algorithmic}
\end{algorithm}

\begin{figure*}[t]
    \centering
    \includegraphics[width=0.8\linewidth, keepaspectratio]{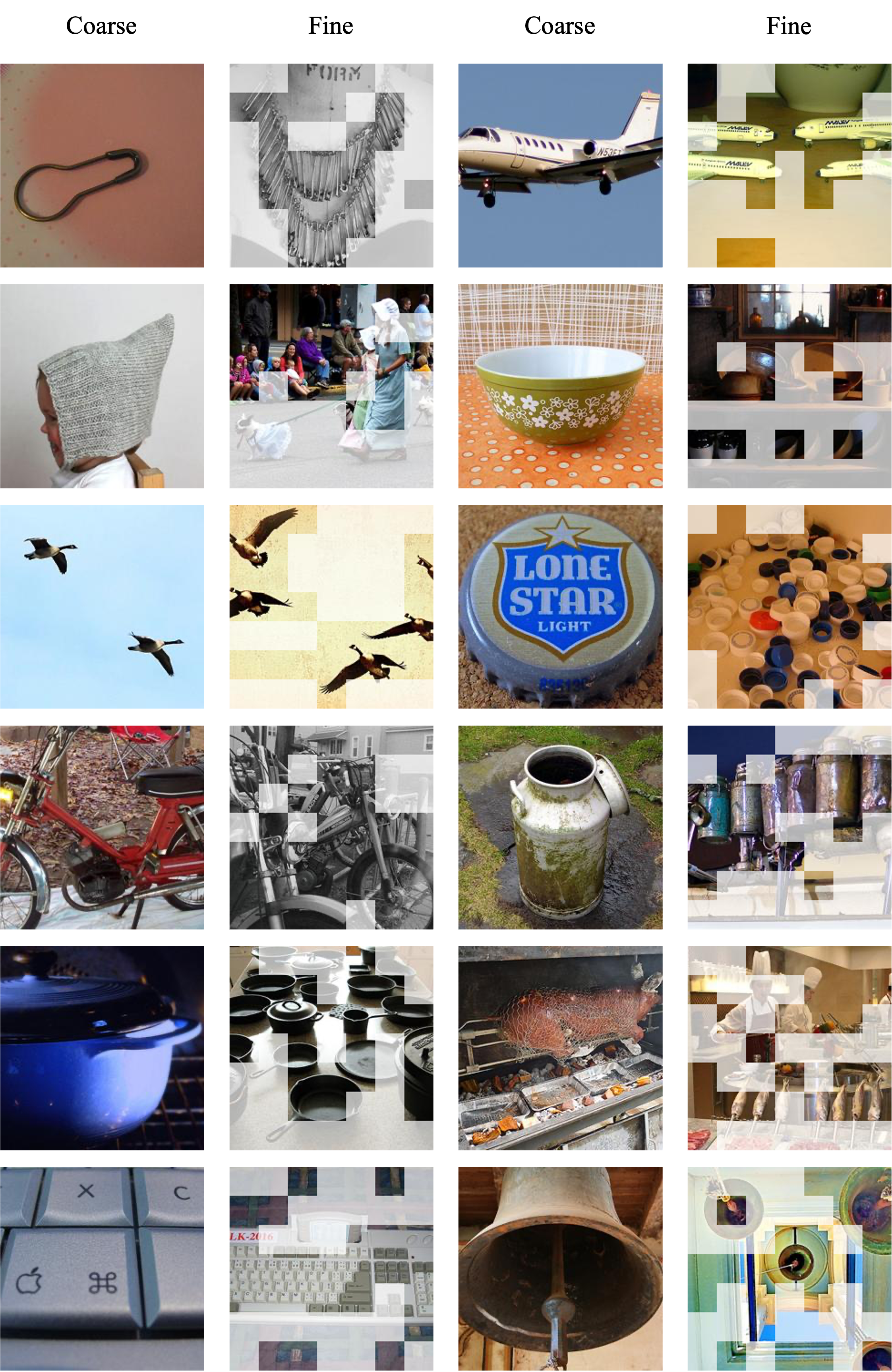}
    \caption{Visualization results of images that pass through only the coarse inference stage (\textbf{Coarse}) and through both coarse and fine stages (\textbf{Fine}).}
    \label{fig:cfcompare}
\end{figure*}

\end{document}


\maketitlesupplementary

\definecolor{mygreen}{RGB}{89,144,144}


\section{Implementation Details}
Our experiments span three major vision tasks: ImageNet-1K classification, ADE20K semantic segmentation, and COCO object detection and instance segmentation. For ImageNet-1K, we follow the standard ViM training configuration and optimize models on $224\times224$ inputs using AdamW (momentum 0.9, weight decay 0.05, batch size 256) with a cosine decay schedule from an initial learning rate of $1\times10^{-3}$ and EMA over 300 epochs. Training employs common augmentation strategies such as random cropping, horizontal flipping, Mixup, label smoothing, and random erasing. The first 200 epochs treat all patches as fine tokens, while the final 100 epochs activate informative patch selection using the hyperparameter~$\alpha$; at inference, we apply center cropping and evaluate under different routing thresholds~$\eta$. This setup ensures a consistent comparison with existing ViM-based token optimization methods and provides controlled evaluation of adaptive token usage.

For semantic segmentation on ADE20K, we adopt UperNet and train for 160K iterations using AdamW (weight decay 0.01, batch size 16) with 1500-step linear warmup and subsequent decay, using $512\times512$ inputs. To support early-exit inference, an auxiliary image-level classification head is incorporated during training, supervised with multi-label targets extracted from the segmentation masks. This auxiliary supervision enhances global semantic awareness without altering the standard segmentation framework. Data augmentation includes random flipping, scale jittering within [0.5, 2.0], and photometric distortions, and evaluation resizes images so that the shorter side is 512 pixels.

For COCO detection and instance segmentation, we use Cascade Mask R-CNN and train for 380K iterations using AdamW (weight decay 0.1, batch size 64). Large-scale jittering is applied to produce $1024\times1024$ inputs, and images are resized during evaluation to a 1024-pixel short side. These settings follow widely adopted detection protocols, ensuring that the improvements brought by our adaptive design are isolated from confounding factors related to training schedules or augmentation choices.

\section{More Visualized Results}

\begin{figure*}[t]
    \centering
    \includegraphics[width=\linewidth,height=\textheight, keepaspectratio]{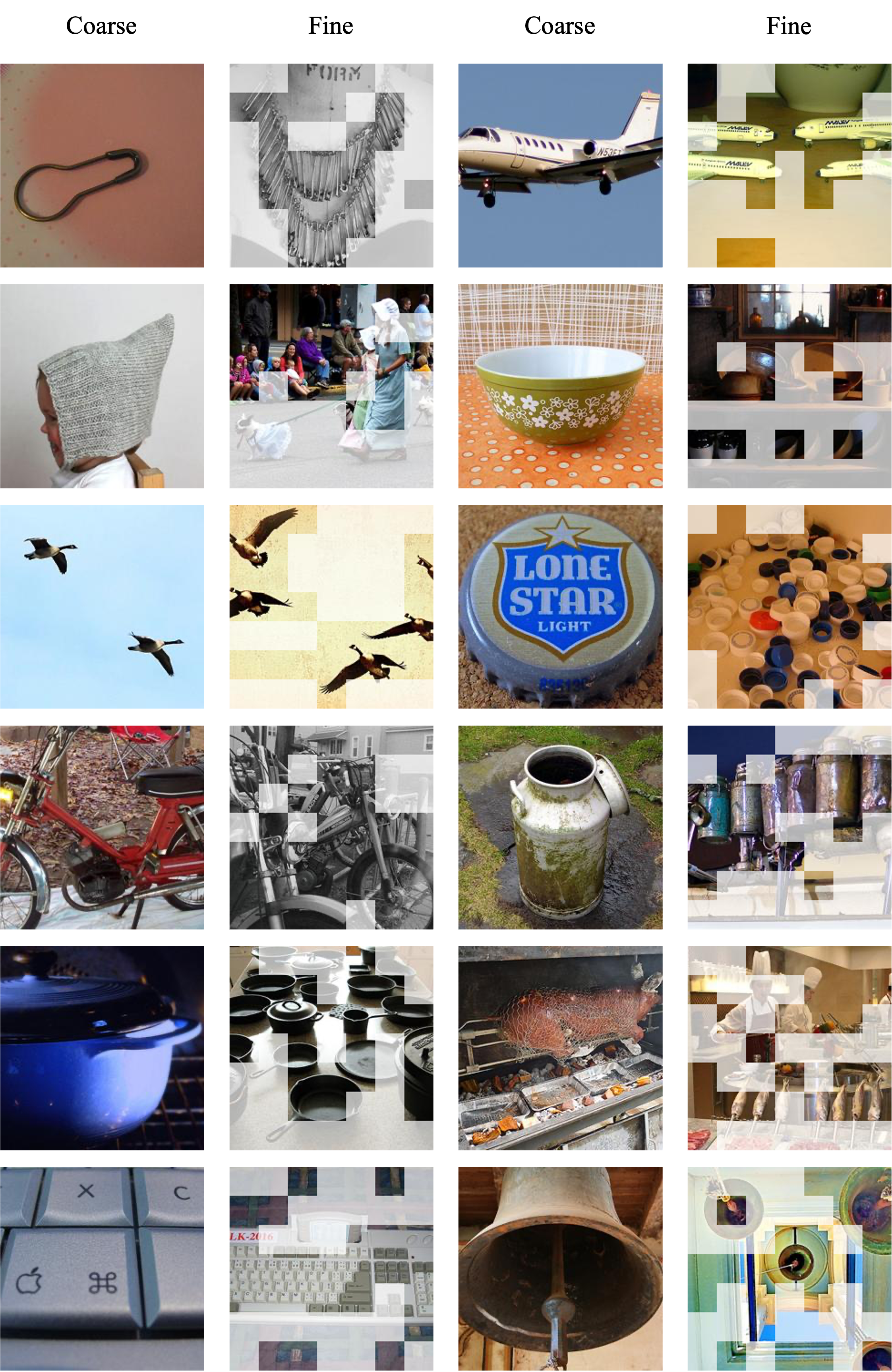}
    \caption{Visualization results of images that pass through only the coarse inference stage (\textbf{Coarse}) and through both coarse and fine stages (\textbf{Fine}).}
    \label{fig:cfcompare}
\end{figure*}

Figure~\ref{fig:cfcompare} presents additional visualizations illustrating the effectiveness of our method. While ViM extracts general visual features for classification, MambaScope selectively refines semantically important coarse tokens into fine-grained representations. This behavior enables the model to focus on critical areas, improving both efficiency and accuracy. The results validate the effectiveness of our coarse-to-fine token processing strategy.

\begin{algorithm}[t]
\small
\caption{Feature Reuse from Coarse to Fine Stage}
\label{alg:feature_reuse}
\begin{algorithmic}[1]
\State \textbf{Input:} coarse features $z_c$, fine features $x$
\State \textbf{Output:} fused features $x_f$
\State \textbf{Notation:} $B$ = batch size, $M$ = token length, $C$ = channel dimension

\State \textcolor{mygreen}{\textit{\# Fine-stage patch embedding and CLS insertion}}
\State $x = \text{patch\_embed}(x)$
\State $B, M, C = x.\text{shape}$
\State $cls\_token = \text{expand}(cls\_token, B, 1, C)$
\State $x = \text{concat}(x[:, :M//2, :], cls\_token, x[:, M//2:, :])$
\Statex
\State \textcolor{mygreen}{\textit{\# Transform coarse features and upsample}}
\State $z_c = \text{reuse\_block}(z_c)$
\State $z_c = \text{interpolate}(\text{reshape}(\text{transpose}(z_c, 1, 2), B, C, \sqrt{N}, \sqrt{N}), \text{size}=(patch\_h, patch\_w), \text{mode}=\texttt{nearest})$

\State $z_c = \text{transpose}(\text{reshape}(z_c, B, C, M), 1, 2)$
\Statex
\State \textcolor{mygreen}{\textit{\# Zero pad CLS position in reused features and fusion}}
\State $zero = \text{zeros}(B, 1, C)$
\State $z_c = \text{concat}(z_c[:, :M//2, :], zero, z_c[:, M//2:, :])$
\State $x_f = x + z_c$
\end{algorithmic}
\end{algorithm}

\begin{algorithm}[t]
\small
\caption{Informative Patch Selection}
\label{alg:informative_selection}
\begin{algorithmic}[1]
\State \textbf{Input:} Importance scores $A$, coarse tokens $z_c$, fine tokens $x$
\State \textbf{Output:} Selected and fused tokens $x$
\State \textbf{Notation:} $\alpha$: selection ratio, $N$: number of coarse tokens, $B$: batch size, $N'$: number of selected fine-grained tokens

\Statex
\State \textcolor{mygreen}{\textit{\# Sort tokens by importance (excluding CLS)}}
\State $A = \text{remove\_cls}(A)$
\State $k = \lceil \alpha \cdot N \rceil$
\State $policy = \text{argsort}(A, \text{desc=True})$
\State $imp, unimp = \text{sort}(policy[:, :k]), \text{sort}(policy[:, k:])$
\Statex
\State \textcolor{mygreen}{\textit{\# Extract unimportant coarse tokens}}
\State $z_{unimp} = \text{select}(z_c, \text{remap\_cls}(unimp))$
\Statex
\State \textcolor{mygreen}{\textit{\# Convert important coarse indices to fine indices}}
\State $imp\_fine = \text{sort}(\text{reshape}(\text{coarse2fine}(imp), B, -1))$
\State $x_{imp} = \text{select}(x, \text{remap\_cls}(imp\_fine))$
\Statex
\State \textcolor{mygreen}{\textit{\# Insert CLS token and fusion.}}
\State $cls = \text{expand}(cls\_token, B, 1, C)$
\State $m = N' // 2$
\State $x_{imp}^{cls} = \text{concat}(x_{imp}[:, :m, :], cls, x_{imp}[:, m:, :])$
\State $x = \text{concat}(x_{imp}^{cls}, z_{unimp})$
\end{algorithmic}
\end{algorithm}

